\documentclass{jas}

\usepackage{csquotes,tikz,lipsum, fancyhdr}
\usepackage{subfigure}
\usetikzlibrary{shapes}

\newcommand{\rr}{\mathbb{R}}

\newcommand{\mbf}{\mathbf}

\title{Image Processing for Motion Magnification}

\author[1]{N. Egidi}
\author[1, $\star$]{J. Giacomini}
\author[2]{P. Leonesi}
\author[1]{P. Maponi}
\author[2]{F. Mearelli}
\author[1]{E. Trebovi\'{c}}

\affil[1]{University of Camerino}
\affil[1]{School of Science and Technology, Mathematics Division \vspace{0.2cm}}
\affil[2]{2T SYSTEM srl}
\affil[2]{Viale Bigioli 102, 62027, San Severino Marche (MC)}

\Email{josephin.giacomini@unicam.it}
\fancypagestyle{plain}{\fancyfoot{}} 

\Subdate{dd/mm/yyyy}

\Volume{W}
\Issue{X}
\Year{YYYY}
\Artnum{Z}
\DOI{0000.000000}
\Pubdate{DD/MM/YYYY}
\Comby{R. Cavoretto}

\begin{document}
	\maketitle
	\begin{abstract}
        Motion Magnification (MM) is a collection of relative recent techniques within the realm of Image Processing. The human visual field can not capture all possible displacement of an object of interest, in particular the smallest one. This is the main motivation of introducing these techniques, in fact, the goal is to opportunely process a video sequence to obtain as output a new video in which motions are magnified and visible to the viewer. These techniques can amplify very subtle motions, imperceptible to the human eye, on proper video sequences, made by the aid of high resolution and high speed cameras. In this work, we propose some preliminary results on MM, developing a technique called Phase-Based Motion Magnification which is performed in the Fourier Domain and rely on the Fourier Shifting Property. In particular, we show the mathematical motivation at the foundation of this method, focusing on some basic test made up using synthetic images.
	\end{abstract}
	
	\keywords{Image Processing, Motion Magnification, Fourier Analysis.}
	
	\section{Introduction}
        Image Processing, which usually abbreviates \emph{Digital Image Processing} (DIP), consists on an ensemble of methods which enable to manipulate or altering an image to obtain a desired result, typically for improving its visual quality or extracting useful information from it. Image processing requires the use of a variety of techniques and algorithms to modify or analyze images and it is a fundamental component of artificial intelligence, computer vision and many other fields, \cite{GonzalezImage}. In general, an image, can be seen as a two-dimensional function $I(x,y)$, where $x$ and $y$ are the spatial (or plane) coordinates and the amplitude $I$ at any pair of coordinates $(x,y)$ is called intensity or gray level of the image at that point. The manipulation of the image is realized through calculators, for this reason both $x$ and $y$ has to be discrete quantities, and so we refer to the image as a \emph{Digital Image}. Moreover, since it carries information through the two axes, we can intend it as a two dimensional signal, and for this cause DIP can be seen as a specialized form of Digital Signal Processing, where the signal is a two dimensional array rather than a one-dimensional time series. Many of the same mathematical tools are used in both fields, such as the Fourier Transform. The discussion we have done until now can be naturally extended to videos, since they are composed by a sequence of images, also known as \emph{frames}. So, a video, is a three dimensional signal $I(x,y,t)$ since the information varies no solely over the spatial coordinates, but also over time.\\ In the context of DIP, we introduce a technique called Motion Magnification (MM). This is a relatively recent technique, proposed by a research group of the Massachusetts Institute of Technology \cite{MM2005}, \cite{PhaseBased}, \cite{Eulerian}. More specifically, this technique establish its roots in DIP. In fact, given a input video $I(x,y,t)$ it is possible, through appropriate manipulation - by working on the individual frames that make up the video - to produce an output $\Tilde{I}(x,y,t)$. This output video can reveals subtle color changes and, more importantly, movements that would be invisible to the naked eye. The primary goal of MM is the detection of those ``invisible'' movements that are indeed present, and that, when analyzed, can provide significant information about the object, structure, or even person captured in the video. In general, the input video, which may seem static to the observer, can actually contain extremely subtle movements. Objects in the scene could shift by as little as $\frac{1}{50}$ or $\frac{1}{100}$ of a pixel. Through a opportune procedure, these sub-pixel motions can be amplified, transforming them into more pronounced displacements that stretch across multiple pixels, making them visible. Essentially, we can think of MM as a kind of ``microscope'' that, instead of zooming in to reveal more detailed visuals, actually ``zoom in'' on movements or color changes. Imperceptible movements are often much faster than what the human eye can detect, which explains why it is necessary to use appropriate video cameras. These cameras offer the possibility of collecting high-density spatial data (high pixel resolution) at high-speed (high frame rate) from a distant scene of interest.\\ MM has a various field of applications. For instance, there are numerous examples of its application in the medical field. These methodologies can be applied for monitoring respiration, specifically aiming to estimate breath rate using non-contact methods, \cite{ALNAJI20161}, \cite{Mattioli}, or to introduce non invasive method based on MM for the study of heart failure, \cite{Abnousi}, \cite{KAYANI2020107290}. Another particular branch of application of these techniques is \emph{vibration analysis}, \cite{VibrationAnalysis}. This has become a widely and powerful tool that allows engineers to study how structures respond to vibrations. This has become a widely and powerful tool that allows engineers to study how structures respond to vibrations. The estimation of the vibration, is usually performed using the classical contact devices, such as accelerometers. Recently, to perform this task, non-contact devices such as video cameras, are used for monitoring vibrating systems, \cite{Chen1}, \cite{CHEN201558}, \cite{Chen2}, \cite{Davis}. The scientific literature about this subject is extensive. For example, \cite{FIORITI2018375} applied motion magnification and image processing techniques to extract the frequency content of vibrations in several ancient structures in Rome and Istanbul. Additionally, \cite{CiveraetAl} conducted a study on Structural Health Monitoring, aiming to identify the exact instant of occurrence for damage or sudden structural changes. Moreover, \cite{Zhao} for rotating fault diagnosis can be mentioned. \\\\  
        In this paper, we propose a preliminary study of the so called Phase-based MM, \cite{PhaseBased}. This approach to MM is performed in the frequency domain, and exploits the Fourier Shift Theorem. This algorithm is effective for global movements due to the Fourier basis, which is defined in the entire domain. The magnification procedure has been implemented in MATLAB and the ``Code metadata'' are:
        \begin{table}[htpb!]
            \centering
            \begin{tabular}{l l}
                Current code version  &  v. 1.0\\
                Permanent link to repository & \url{https://github.com/eTrebo98/MotMagArt1} \\
                Code versioning system used & git \\
                Software code languages & MATLAB 
            \end{tabular}
        \end{table}
        
        This paper is organized as follows.\\ In section 2, we explore the mathematical background under the proposed procedure, describing the steps needed to perform the magnification of a shift between two consecutive frame of a video. In section 3, since the algorithm needs to be applied on a calculator, we focus our attention on the discretized version of it, introducing the Discrete Fourier Transform (DFT). Next, in section 4, we show some numerical results made up with MATLAB using synthetic gray-scale images. We put much emphasis on presenting the amplification of the movement between two consecutive frames, furthermore we consider also an example of an output video sequence in which the shift are exaggerated respect to the input one. In section 5, we provide conclusions and future developments.
        
	\section{Fourier Analysis and Motion Magnification}
        Many Image Processing techniques are based on the Fourier Transform. In fact, many task such filtering, or noise reduction are performed in the frequency domain \cite{GonzalezImage}. Another important aspect in which spectral techniques can be useful, is the motion estimation in video-sequences. There are several methods based on the Optical Flow Equation that are effective in performing this task \cite{HornShunck}, \cite{LucasKanade}. One of these due to \cite{Fleet} is specialized for the computation of 2D component velocity from image sequences, by analyzing local phase changes. Motion causes shifts in the phase and, by measuring how the phase evolves in small regions, the procedure compute the object’s speed. So, the motivation for the proposed motion magnification procedure lies on the properties of the phase in image sequences, particularly its sensitivity to small motions and robustness to changes in intensity. This method uses the phase information from the frequency domain, which provides a more stable and precise representation of motion, especially for subtle, or sub-pixel movements.
        \subsection{Fourier Analysis}
        Briefly, we recall the definition of the $d$-dimensional Fourier transform.
        \begin{defn}
        Let $f \in L^1(\rr^d)$. The Fourier transform $\mathcal{F} f = \hat{f}$ of $f$ is defined by:
			\begin{equation}
				\label{FourierTransformR2}
				\mathcal{F}f (\mbf{\omega}) = \hat{f}(\mbf{\omega}) = \int_{\rr^d}f(\mbf{x})e^{-2 \pi \iota\mbf{\omega}^T\cdot\mbf{x}} \,d\mbf{x}, \quad \mbf{\omega} \in \rr^d,
			\end{equation}
        where $\mbf{x} = (x_1, x_2, \cdots, x_d)$, $\mbf{\omega} = (\omega_1, \omega_2, \cdots, \omega_d)$, and $\cdot$ denotes the Euclidean inner product.
	\end{defn}
       The Fourier Transform, satisfies many important properties \cite{FourierStade}, in this paper we recall one of these, since it is fundamental for the magnification algorithm.
       \begin{thm}[Fourier Shift Theorem]
		\label{ShiftTheorem}
		Let $\mbf{\delta} \in \rr^d$, then:
		\begin{equation}
				\label{ShiftingPropR2}
				\mathcal{F}[f(\mbf{x} -\mbf{\delta})] = e^{-2\pi \iota \mbf{\delta}^T \cdot \mbf{\omega}}\hat{f}(\mbf{\omega}).
		\end{equation}
	\end{thm}
        \begin{proof}
           See \cite{FourierStade} for details.
        \end{proof}
        The last result says that if a given function is shifted in the positive direction by a vector $\mbf{\delta}$, no Fourier components changes in amplitude. Therefore, it is expected that the changes in its Fourier Transform will be confined to the phase. In other words, a translation in the spatial domain corresponds to a multiplication by a complex exponential in the frequency domain.\\
        Moreover, for a real function $f(\mbf{x})$ we have the so called hermitian-symmetric property:
	\begin{equation*}
		\hat{f}(-\mbf{\omega}) = \overline{\hat{f}(\mbf{\omega})}, \quad \forall \mbf{\omega} \in \rr^d,
	\end{equation*}
	where $\overline{\cdot}$ denotes the complex conjugate.\\
       Clearly, it is important to remind that, once performed the operations in the Fourier Domain, under appropriate condition, it is possible to return back, in the original domain, by taking the Inverse Fourier Transform, that is:
       \begin{thm}
			If $f, \hat{f} \in L^1(\rr^d)$ then
			\begin{equation}
				\label{FourierInvR2}
				f(\mbf{x}) = \int_{\rr^d}\hat{f}(\mbf{\omega})e^{2 \pi \iota\mbf{\omega}^T\cdot\mbf{x}}\, d\mbf{\omega}\,
			\end{equation}
			for almost all $\mbf{x}\in \rr^d$. In addition, if $f \in C^0(\rr^d)$ then \eqref{FourierInvR2} is true of all $\mbf{x}\in \rr^d$.
		\end{thm}
  \begin{proof}
      See \cite{FourierStade} for a detailed proof.
  \end{proof}
      In the following, for our purpose, $d = 2$ since we work with images. Furthermore, we denote with $I(x,y)$ a continuous valued real image, while with $J(n,m)$ the corresponding digital one, obtained by a process of sampling and quantization from $I(x,y)$, \cite{GonzalezImage}. More in detail, $J(n,m)$ is a matrix, such that $n = 0,1,2,\cdots, N-1$, where $N$ is the height of the digital image and $m = 0,1,2,\cdots, M-1$, where $M$ is the width.
        \subsection{Motion Magnification}
       Let us suppose that $I_1(\mbf{x})$ is an image at time $t = 0$, that is the initial frame of a video. The next frame can be viewed as the initial frame which undergoes a small motion $\mbf{\delta} = (\delta_1, \delta_2) \in \rr^2$. That is:
	\begin{equation} \label{ShiftI2Vid}
		I_2(\mbf{x}) = I_1(\mbf{x}+\mbf{\delta}).
	\end{equation}
		The goal is to synthesize a new frame with modified motion, that is
       \begin{equation*}
			\Tilde{I_2}(\mbf{x}) = I_1(\mbf{x}+(1+\alpha)\mbf{\delta}) = I_1(x_1+(1+\alpha)\delta_1,\, x_2 + (1+\alpha)\delta_2).
       \end{equation*}
       Recalling definition \eqref{FourierTransformR2}, we have that the Fourier Transform of the two images are defined as:
		\begin{equation}
                \label{FTI2}
			\hat{I}_j(\mbf{\omega}) = \int_{-\infty}^{\infty}I_j(x,y) e^{-2\pi \iota \mbf{\omega}^T\cdot\mbf{x}}\, d\mbf{x}, \quad \mbf{\omega} \in \rr^2, \quad j = 1,2.
		\end{equation}
		Since $I_2(\mbf{x})$ is defined like \eqref{ShiftI2Vid}, we can rewrite \eqref{FTI2} by using the Shifting property \eqref{ShiftingPropR2}. So we have:
		\begin{equation}
			\label{FTShiftI2}
			\hat{I}_2(\mbf{\omega}) = e^{2\pi \iota \mbf{\delta}^T \cdot \mbf{\omega}}\hat{I}_1(\mbf{\omega}).
		\end{equation}
		Since, for all $\mbf{\omega} \in \rr^2$, $\hat{I}_j(\mbf{\omega})$, $j = 1,2$ are complex numbers, we can express them as:
		\begin{equation}
                \label{PolarFormI2Vid}
			\hat{I}_j(\mbf{\omega}) = |\hat{I}_j(\mbf{\omega})|e^{\phi_j(\mbf{\omega})}, \quad j = 1,2.
		\end{equation}
		At this stage, we can define, for all $\mbf{\omega} \in \rr^2$ the quantity:
		\begin{equation}
			\label{EQuantity}
			E(\mbf{\omega}) = \dfrac{\hat{I}_2(\mbf{\omega})}{\hat{I}_1(\mbf{\omega})} = \dfrac{|\hat{I}_2(\mbf{\omega})|e^{\phi_2(\mbf{\omega})}}{|\hat{I}_1(\mbf{\omega})|e^{\phi_1(\mbf{\omega})}} = \dfrac{|\hat{I}_1(\mbf{\omega})|e^{\phi_1(\mbf{\omega})}e^{2\pi \iota \mbf{\delta}^T \cdot \mbf{\omega}} }{|\hat{I}_1(\mbf{\omega})|e^{\phi_1(\mbf{\omega})}} = e^{2\pi \iota \mbf{\delta}^T \cdot \mbf{\omega}}.
		\end{equation}
		where the third equality of \eqref{EQuantity} is due to Relation \eqref{FTShiftI2}. Now, by multiplying the Fourier Transform $\hat{I}_2(\mbf{\omega})$ for $E(\mbf{\omega})^{\alpha}$, we obtain:
		\begin{equation*}
			\widehat{\Tilde{I_2}}(\mbf{\omega}) = E(\mbf{\omega})^{\alpha}\hat{I}_2(\mbf{\omega}) = e^{2\pi \iota (\mbf{\omega} + (1+\alpha)\mbf{\delta})}\hat{I_1}(\mbf{\omega}).
		\end{equation*}
		Finally taking the Inverse Fourier transform \eqref{FourierInvR2} of $\widehat{\Tilde{I_2}}(\mbf{\omega})$, we have:
		\begin{equation*}
			\Tilde{I_2}(\mbf{x}) = I_1(\mbf{x}+(1+\alpha)\mbf{\delta}),
		\end{equation*}
		which is the magnified frame. This the mathematical motivation under the concept of motion magnification.\\ From a practical point of view, we need to introduce the Discrete Fourier Transform (DFT) in order to compute the magnification procedure on the calculator.
       
        \section{DFT and Motion Magnification}
       In this section, we introduce the basic definition of the DFT and some key properties useful in the sequel, and then we discuss the motion magnification algorithm.
		\begin{defn}
			Let $J(n,m)$, with $n = 0,1\cdots, N-1$ and $m = 0,1,\cdots, M-1$, a digital image of size $N \times M$. The 2D DFT of $J$ denoted by $\hat{J}$, is given by:
			\begin{equation}
				\label{DFT2}
				\hat{J}(k,l) = \sum_{n = 0}^{N-1}\sum_{m = 0}^{M-1}J(n,m)W_N^{kn} W_M^{lm},
			\end{equation}
			for $k = 0,\cdots, N-1$ and $l = 0,\cdots, M-1$.\\
			The 2D Inverse DFT (IDFT) of $\hat{J}$ is defined as:
			\begin{equation}
				\label{IDFT2}
				J(n,m) = \dfrac{1}{NM}\sum_{k = 0}^{N-1}\sum_{l = 0}^{M-1}\hat{J}(k,l)W_N^{-kn} W_M^{-lm},
			\end{equation}
			with $n = 0,\cdots, N-1$ and $m = 0,\cdots, M-1$.
                Where $W_N^{kn} = e^{-2 \pi \iota \frac{kn}{N}}$ and $W_M^{lm} = e^{-2\pi \iota \frac{lm}{M}}$.
		\end{defn}
        It can be proved that the Discrete Fourier Transform satisfies the following translation properties, as in theorem \eqref{ShiftTheorem}:
		\begin{prop}
			If $J(n,m)$ has Fourier transform $\hat{J}(k,l)$ then $J(n+\delta_1, m+\delta_2)$ has Fourier Transform given by:
			\begin{equation}
                \label{TranslationDFT2}
				\hat{J}(k,l)e^{2\pi \iota(\frac{k\delta_1}{N} + \frac{l\delta_2}{M})}.
			\end{equation}
		\end{prop}
           \begin{proof}
               See \cite{OppenheimDigital} for more details.
           \end{proof}
		That is, multiplying $\hat{J}(k,l)$ by the such complex exponential shifts the origin of $J(n,m)$ to $(\delta_1,\delta_2)$.
        Moreover, by the quantities $W_N^{kn}$, we have that the DFT of a real function is hermitian symmetric, which implies it has the following symmetries about the center of the matrix $J(n,m)$:
		\begin{equation}
                \label{DFTSym}
		    \begin{split}
                    \hat{J}(N-k,M-l) = \overline{\hat{J}(k,l)},\quad k = 0,\cdots, N-1 \quad l = 0,\cdots, M-1\\
                    \hat{J}(k,M-l) = \overline{\hat{J}(N-k,l)}\quad k = 0,\cdots, N-1 \quad l = 0,\cdots, M-1.
		    \end{split}
		\end{equation}
  
        \subsection{Algorithm implementation} \label{sec:Implementation}
        We discuss the algorithmic approach used to perform the magnification. Let suppose that $J_j(n,m)$, $j = 1,2$ are two consecutive digital frame of a video sequence. In particular, $J_2(n,m) = J_1(n+\delta_1, m + \delta_2)$, with $n = 0,1,\cdots, N-1$, $m = 0,1,\cdots, M-1$ and $\mbf{\delta} = (\delta_1, \delta_2) \in \rr^2$. To compute the magnification by a factor $\alpha$, with $\alpha \in \rr$, we consider the DFT of the two frames, that is, by definition \eqref{DFT2}:
		\begin{equation}
                \label{DFT2I2}
			\hat{J}_j(k,l) = \sum_{n = 0}^{N-1}\sum_{m = 0}^{M-1}J_1(n,m)W_N^{kn} W_M^{lm}, \quad j = 1,2.
		\end{equation}
		where $k=0,1,\cdots,N-1$ and $l = 0,1,\cdots,M-1$.
		But, by relation \eqref{TranslationDFT2}, we can rewrite \eqref{DFT2I2} as follows:
		\begin{equation*}
			\hat{J}_2(k,l) = e^{2\pi \iota (\delta_1 \frac{k}{N} + \delta_2 \frac{l}{M})} \hat{J}_1(k,l).
		\end{equation*}
	Now, we proceed by constructing the magnified signal. First of all, we consider the quantity
   \begin{equation}
	\label{Ratio}
	E(k,l) = \dfrac{\hat{J}_2(k,l)/ |\hat{J}_2(k,l)|}{\hat{J}_1(k,l)/ 
        |\hat{J}_1(k,l)|} = e^{2\pi \iota (\delta_1 \frac{k}{N} + \delta_2 \frac{l}{M})},
   \end{equation}
   where we divide both $\hat{J}_2(k,l)$ and $\hat{J}_1(k,l)$ by their modulus respectively, to guarantee that $|E(k,l)| = 1$.\\ 
   In order to compute the DFT of the magnified frame $\widehat{\Tilde{J_2}}(k,l)$ we have to notice that by \eqref{DFTSym} there is a hermitian symmetry with respect to the central element of the matrix which constitutes the DFT of the frame, thus we compute the DFT of the magnified frame not for all $k$ and $l$. In particular, it is necessary to consider different cases, depending on the parity of $N$ and $M$.\\ 
   \begin{itemize}
   \item $N$ and $M$ are both odd.
    \begin{itemize}
        \item Symmetry along the first row, $k = 0$. The symmetry occurs only along the row itself, that is, for $l = 1, \cdots, \frac{M-1}{2}$, we compute:
		\begin{equation*}
			\widehat{\Tilde{J_2}}(0,l) = \hat{J}_2(0,l) (E(0,l))^{\alpha} \quad \text{and} \quad \widehat{\Tilde{J_2}}(0,M-l) = \overline{\widehat{\Tilde{J_2}}(0,l)}.
		\end{equation*}
        \item Symmetry along the first column, $l = 0$. The symmetry is only along the column itself, so for $k = 1, \cdots, \frac{N-1}{2}$, we compute:
        \begin{equation*}
			\widehat{J_2^{mag}}(k,0) = \hat{J}_2(k,0) (E(k,0))^{\alpha} \quad \text{and} \quad \widehat{J_2^{mag}}(N-k,0) = \overline{\widehat{J_2^{mag}}(k,0)}.
	\end{equation*}
       \item In the limit case for which both $k = 0$ and $l = 0$ there is no conjugate symmetric to compute.
       \item For the rest of the DFT array, symmetry is governed by the general two-dimensional conjugate symmetry. So, For all $k = 1, \cdots, \frac{N-1}{2}$, $l = 1, \cdots, \frac{M-1}{2}$ we calculate: 
       \begin{equation*}
			\widehat{\Tilde{J_2}}(k,l) = \hat{J}_2(k,l) (E(k,l))^{\alpha} \quad\text{and}\quad \widehat{\Tilde{J_2}}(N-k,M-l) = \overline{\widehat{\Tilde{J_2}}(k,l)}.
		\end{equation*}
		Also $\widehat{\Tilde{J_2}}(k,M-l)$ needs to be explicitly computed, that is: 
	\begin{equation*}
			\widehat{\Tilde{J_2}}(k,M-l) = \hat{J}_2(k,l) (E(k,M-l))^{\alpha} \quad\text{and}\quad \widehat{\Tilde{J_2}}(N-k,l) = \overline{\widehat{\Tilde{J_2}}(k,M-l)}.
        \end{equation*}  
    \end{itemize}
    \item $N$ and $M$ are both even.
    \begin{itemize}
        \item An additional symmetry appear along the center row, at $k = \frac{N}{2}$. In particular, for $l= 1,2, \cdots, \big(\frac{M}{2} -1\big)$ we compute:
		\begin{equation*}
			\widehat{\Tilde{J_2}}(\tfrac{N}{2},l) = \hat{J}_2(\tfrac{N}{2},l) (E(\tfrac{N}{2},l))^{\alpha}\quad \text{and} \quad \widehat{\Tilde{J_2}}(\tfrac{N}{2},M-l) = \overline{\widehat{\Tilde{J_2}}(\tfrac{N}{2},l)}.
		\end{equation*}
       \item An additional symmetry appear along the center column, at $l = \frac{M}{2}$. In particular, for $k =1, 2,\cdots,\big(\frac{N}{2}-1\big)$ we compute:
       \begin{equation*}
			\widehat{\Tilde{J_2}}(k,\tfrac{M}{2}) = \hat{J}_2(k,\tfrac{M}{2}) (E(k,\tfrac{M}{2}))^{\alpha}\quad \text{and} \quad \widehat{\Tilde{J_2}}(N-k,\tfrac{M}{2}) = \overline{\widehat{\Tilde{J_2}}(k,\tfrac{M}{2})}.
		\end{equation*}
       \item The points at the intersections, such as $\widehat{\Tilde{J_2}}(0,\tfrac{M}{2})$, $\widehat{\Tilde{J_2}}(\tfrac{N}{2},0)$, $\widehat{\Tilde{J_2}}(\tfrac{N}{2},\tfrac{M}{2})$, are purely real.
       \item The remaining regions in the DFT are computed as in the previous case.
    \end{itemize}
    \item The cases with $N$ even and $M$ odd, or vice versa, are handled as an appropriate combination of the previously discussed situations.
    \end{itemize}
    After the procedure has been completed, we recover the magnified frame by taking its inverse DFT \eqref{IDFT2}, that is:
		\begin{equation*}
			\Tilde{J_2}(n,m) = \dfrac{1}{NM} \sum_{k=0}^{N-1}\sum_{l=0}^{M-1}\widehat{\Tilde{J_2}}(k,l) W_N^{-kn}W_M^{-lm},
		\end{equation*}
		where $n = 0,\cdots,N-1$ and $m = 0, \cdots, M-1$.\\From a practical point of view the two initial DFTs and the final IDFT are computed using a FFT algorithm for 2-dimensional arrays, \cite{FFT2}.
    
	\subsection{MATLAB code listings} \label{sec:code}
        The algorithmic procedure discussed in section \ref{sec:Implementation} has been implemented in MATLAB. Here we illustrate the function that computes the magnified frame for odd-sized images. A generalized version of the function for any image size can be found on \url{https://github.com/eTrebo98/MotMagArt1}.\\
        \begin{itemize}
            \item \textbf{Syntax.} \emph{[imMag, ifail] = CompMagMatrix(im1,im2,alpha)}
            \item \textbf{Purpose.} Computes the digital image containing the magnified motion.
            \item \textbf{Description.} \emph{[imMag, ifail] = CompMagMatrix(im1,im2,alpha)}, takes two gray-scale images $im1$ and $im2$, and computes the magnified digital image $imMag$ with a magnification factor $alpha$ by the procedure outlined in section \ref{sec:Implementation}.
            \item \textbf{Parameters.}
            \begin{itemize}
                \item \textbf{input} $im1$, $im2$ - gray-scale images of data type uint8 with values in the interval $[0,255]$.
                \item \textbf{input} $alpha$ - double scalar specifying the magnification factor of the motion
                \item \textbf{output} $imMag$ - uint8 matrix of the same size as $im1$ and $im2$ representing the digital image with the magnified motion.
                \item \textbf{output} $ifail$ - integer scalar. $ifail = 0$ unless the function detects an error (see Error Indicators and Warnings)
            \end{itemize}
            \item \textbf{Error Indicators and Warnings.} Here is the list of errors or warnings detected by the function:
            \begin{itemize}
                \item $ifail = 1$ - if the size of $im1$ and $im2$ differs.
                \item $infail = 2$ - if $im1$ and $im2$ are not odd-sized.
            \end{itemize}
        \end{itemize}
        In code \ref{code:1} the MATLAB implementation of \emph{[imMag, ifail] = CompMagMatrix(im1,im2,alpha)}.
	
\begin{lstlisting}[language=matlab, caption=MATLAB Function for computing the magnified frame\label{code:1}]
function [imMag, ifail] = CompMagMatrix(im1,im2,alpha)

    %This function perform a magnification between two images (matricies) which differs
    %from each other by a certain amount of motion. The procedure is
    %constructed in such way to work with odd-sized images.
    %   INPUT: im1: first frame in greyscale
    %          im2: second frame shifted, in greyscale
    %          alpha: magnification factor
    %   OUTPUT:imMag: Magnified frame
    %          ifail: integer scalar, ifail = 0 unless the function detects
    %                 an error (ifail = 1 the two images have different
    %                 sizes, ifail = 2 if one N or M are not odd)
    
    %calculate the sizes of the two frames
    ifail = 0;

    [N_1,M_1] = size(im1);
    [N_2,M_2] = size(im2);
    
    %if the frames have different sizes we can't compute this task.
    if N_1 ~= N_2 || M_1 ~= M_2
        ifail = 1;
        imMag = 0;
        return;
    end

    %if one of N or M is even we can't compute this task.
    if mod(N_1,2) == 0 || mod(M_1,2) == 0
        ifail = 2;
        imMag = 0;
        return;
    end

    
    %initialize the dimension of the image
    N = N_1;
    M = M_1;

    fftImMag = zeros(N,M); %matrix which contains the DFT of the magnified frame

    %compute the Discrete Fourier Transform of the two frames. Converting
    %to double the images since they are expressed in greyscale (values from 0
    %to 255).
    fftIm1 = fft2(im2double(im1));
    fftIm2 = fft2(im2double(im2));

    %the Cycle is performed only for the half-size of the matrix, because of
    %the hermitian symmetry of the DFT.
    for k=1:ceil(N/2)
        for l=1:ceil(M/2)
            if (k == 1) %we have a symmetry in the first row of the matrix
                if (l == 1) %the first element of the matrix has no symmetric
                    %calling ComputeMag function, which computes the DFT of the
                    %current (k,l) element of the magnified DFT
                    [fftImMag(k,l)] = CompElemMag(fftIm1(k,l),fftIm2(k,l),alpha);
                else
                    refl = M-l+2;%take the reflected index for the symmetry
                    [fftImMag(k,l), fftImMag(k,refl)] = CompElemMag(fftIm1(k,l),fftIm2(k,l),alpha);
                end
            else
                %Hermitian symmetry along the first column
                if (l == 1)
                    refk = N-k+2; %take the reflected index for the symmetry
                    [fftImMag(k,l), fftImMag(refk,l)] = CompElemMag(fftIm1(k,l),fftIm2(k,l),alpha);
    
                else %now the symmetry is respect the centre of the matrix
    
                    %computing the reflected indices for the symmetry
                    refk = N-k+2;
                    refl = M-l+2;
    
                    %calling the ComputeMag function, which performs the
                    %magnification of the (k,l) element of the matrix and its
                    %conjugate symmetric
                    [fftImMag(k,l), fftImMag(refk,refl)] = CompElemMag(fftIm1(k,l),fftIm2(k,l),alpha);
    
                    %compute the magnification of the (k,refl) element and its
                    %conjugate symmetric.
                    [fftImMag(k,refl), fftImMag(refk,l)] = CompElemMag(fftIm1(k,refl),fftIm2(k,refl),alpha);
                end
            end
        end
    end
    %compute the ifft of the magnified Frame
    imMag = im2uint8(ifft2(fftImMag));
    
end
\end{lstlisting}
        
	
	\section{Numerical Results} \label{sec:math}
        In this section, we present some numerical simulations of the described procedure for motion magnification using test images generated in MATLAB. The goal of these numerical simulations is to validate the theoretical framework discussed for the amplification of movements applied to images. The input frame used in the simulation is a gray-scale image with a circle centered in a specific position of the image. The pixel intensity values vary smoothly from the center of the circle to the edges, representing a transition from white at the center to black at the boundaries. To generate the test image, we used a \emph{sigmoid function} to create a smooth transition at the edges of the circle. This allowed us to avoid hard, pixelated boundaries. The sigmoid function used is defined as follows:
	\begin{equation*}
		S(\rho) = \dfrac{255}{1+e^{\frac{\rho}{\sigma}}},
	\end{equation*}
	where $\rho$ represents the distance from the center of 
        the circle, $\sigma$ is a parameter that controls the color transition at the boundary.
        \subsection{Multi-frame}
\begin{figure}[htpb!]
            \centering
            \subfigure[]{\includegraphics[width=0.3\textwidth]{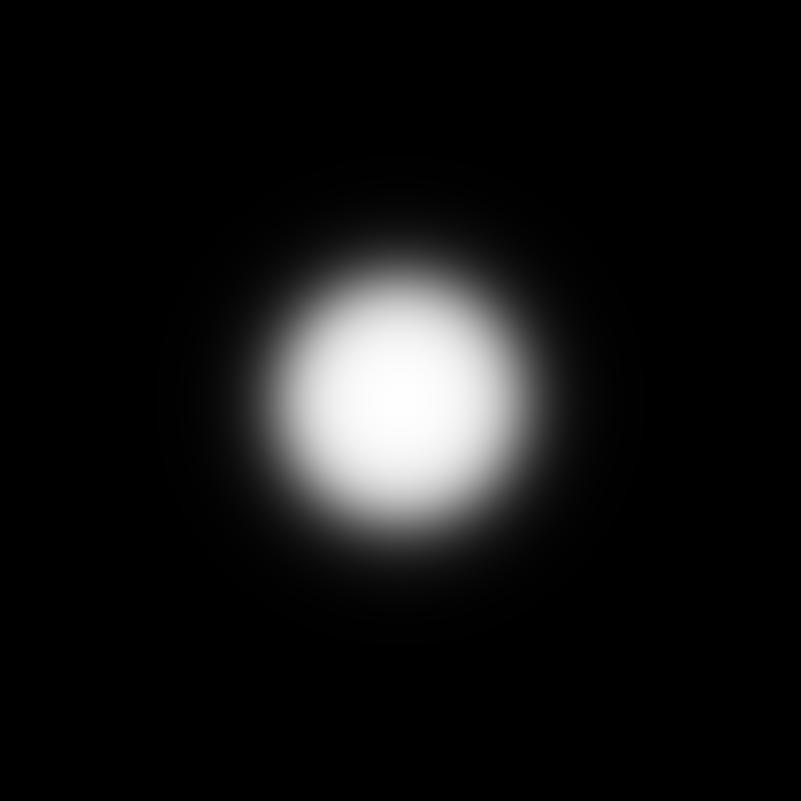}} 
            \subfigure[]{\includegraphics[width=0.3\textwidth]{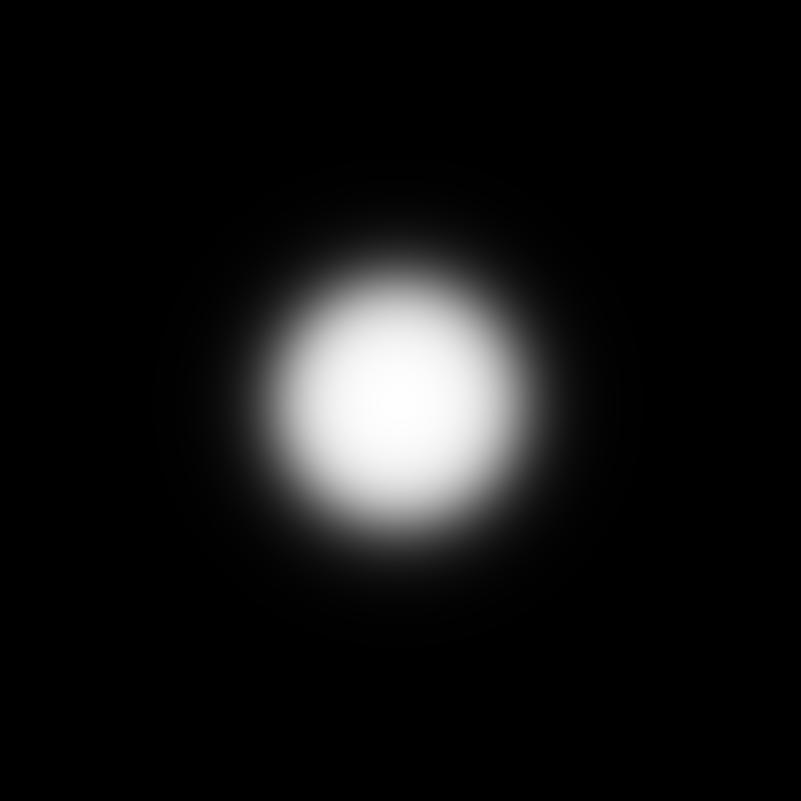}} 
            \subfigure[]{\includegraphics[width=0.3\textwidth]{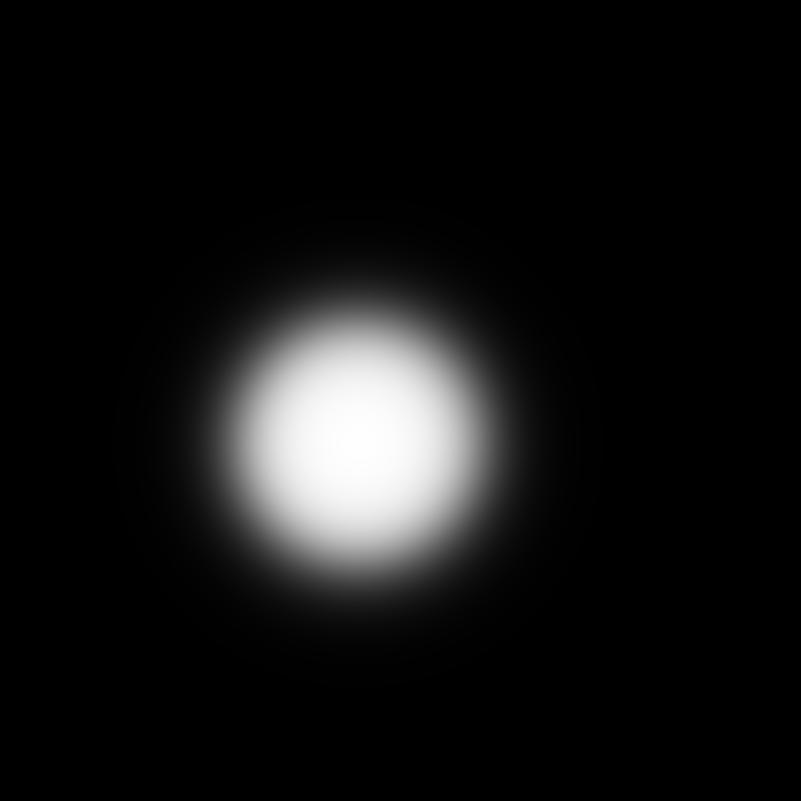}}
    
            \caption{(a) Original Frame (b) Shifted Frame (c) Magnified Frame ($\alpha = 20$)}
            \label{fig:AmpliTwoImag1}
        \end{figure}

\begin{figure}[htpb!]
            \centering
            \subfigure[]{\includegraphics[width=0.3\textwidth]{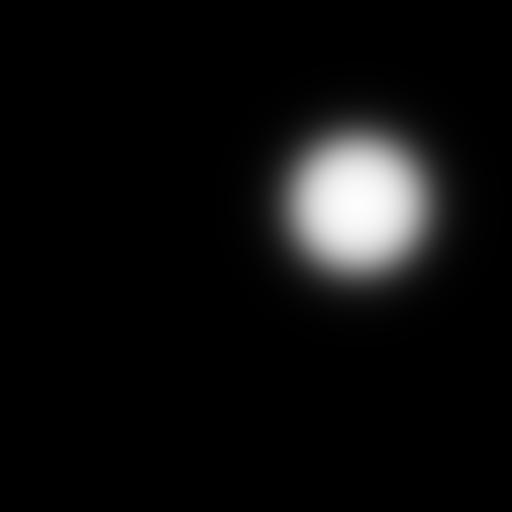}} 
            \subfigure[]{\includegraphics[width=0.3\textwidth]{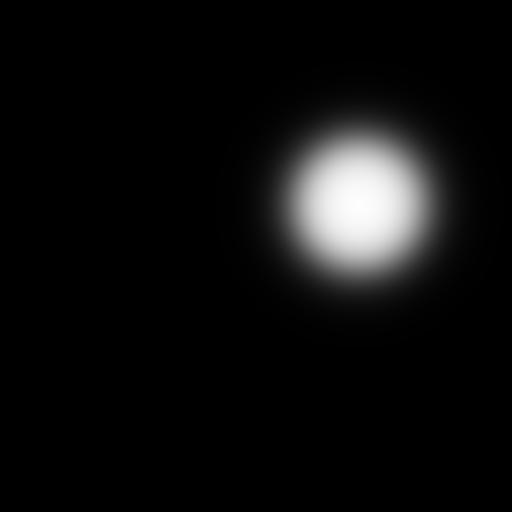}} 
            \subfigure[]{\includegraphics[width=0.3\textwidth]{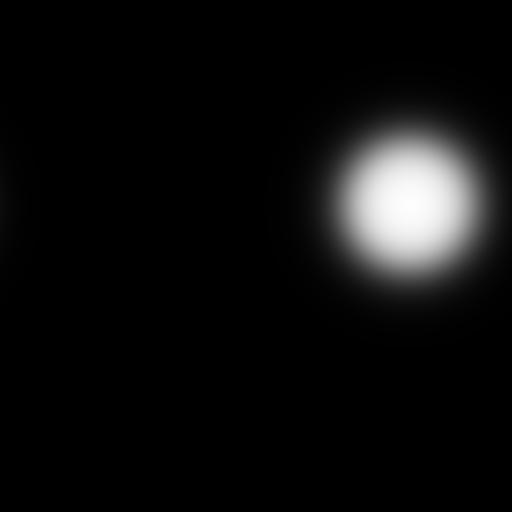}}
    
            \caption{(a) Original Frame (b) Shifted Frame (c) Magnified Frame ($\alpha = 105$)}
            \label{fig:AmpliTwoImag2}
        \end{figure}

\begin{figure}[htpb!]
            \centering
            \subfigure[]{\includegraphics[width=0.3\textwidth]{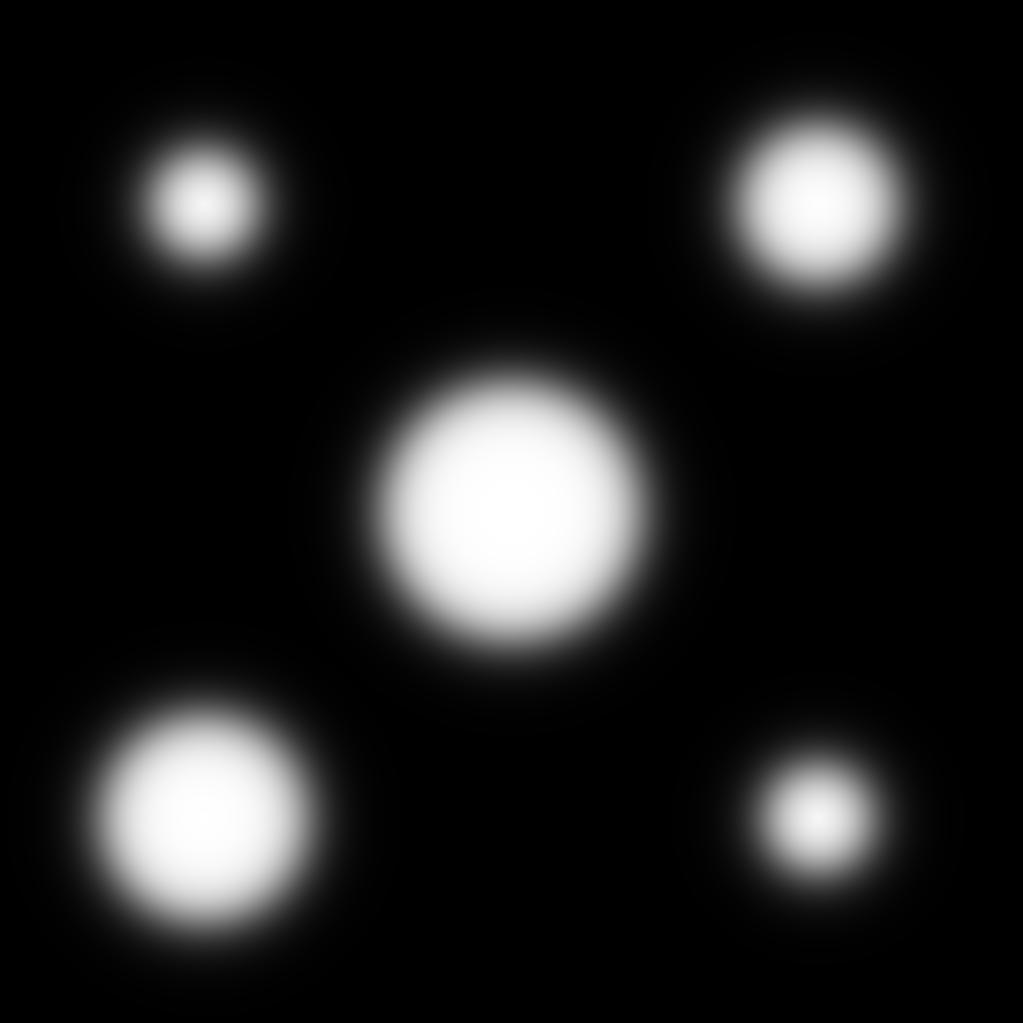}} 
            \subfigure[]{\includegraphics[width=0.3\textwidth]{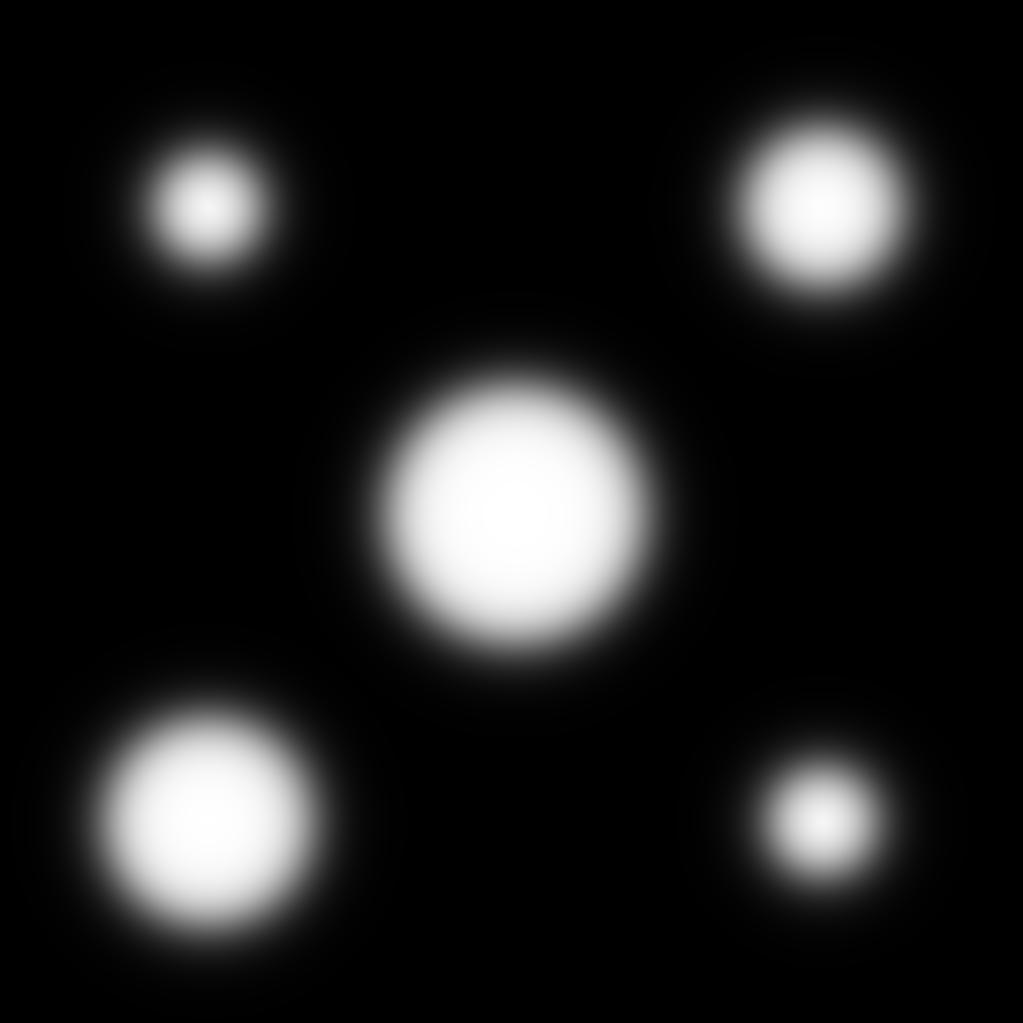}} 
            \subfigure[]{\includegraphics[width=0.3\textwidth]{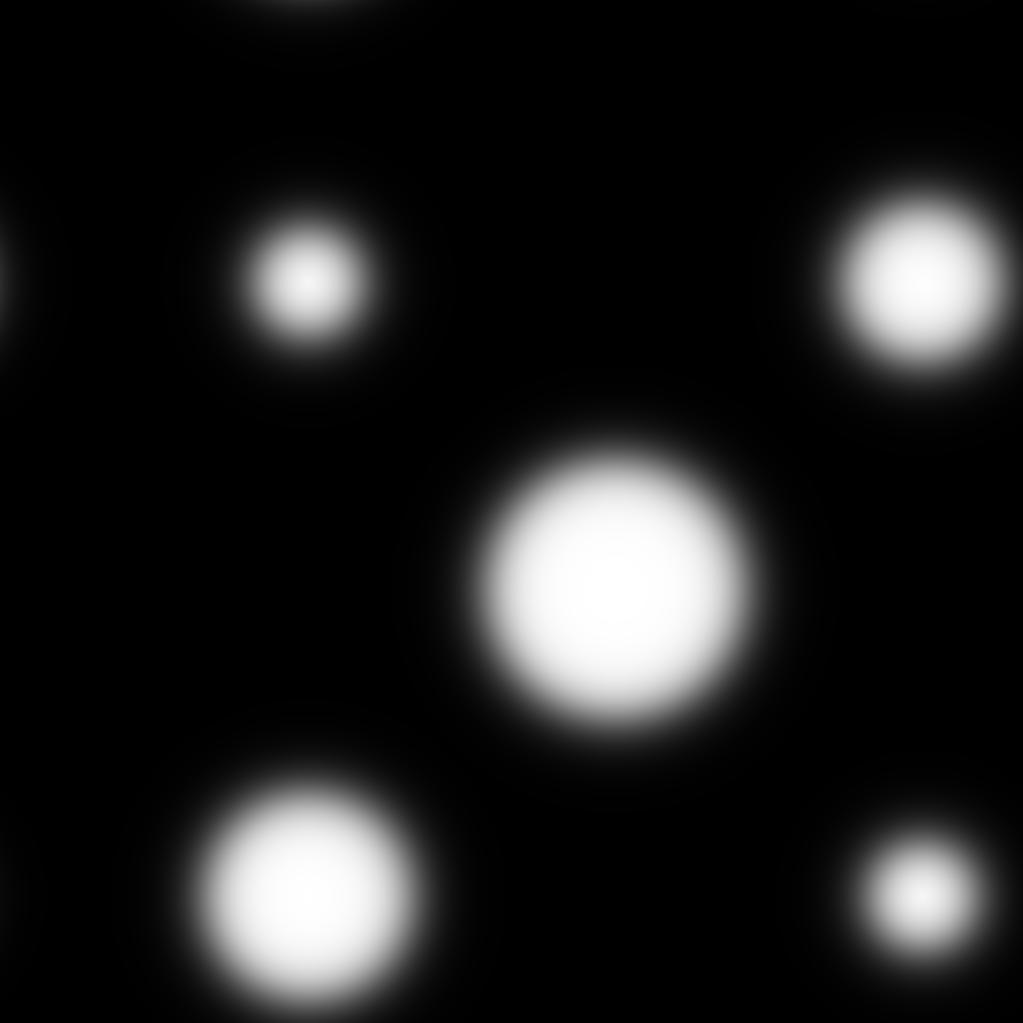}}
    
            \caption{(a) Original Frame (b) Shifted Frame (c) Magnified Frame ($\alpha = 25$)}
            \label{fig:AmpliTwoImag3}
        \end{figure}

        Firstly, to implement the motion amplification, we consider two images, having the same intensity, but the second one has applied a shift of $(\delta_1,\delta_2)$, expressed in pixels.\\ In figure \ref{fig:AmpliTwoImag1}, $\delta_1 = 2$ and $\delta_2 = -2$, that is, all the intensity values of the second image are translated by two pixels down along the rows and two pixels left along the columns. We consider a magnification factor $\alpha = 20$, so that the magnified image is shifted by an amount $(1+\alpha)\delta_1 = 42$ pixels along the rows and $(1+\alpha)\delta_2 = -42$ pixels along the columns. We show the reference frame, the shifted frame and the magnified one. All three images have a resolution of $801 \times 801$ pixels.\\
        In figure \ref{fig:AmpliTwoImag2}, $\delta_1 = 0$ and $\delta_2 = 1$, that is, all the intensity values of the second image are translated by 1 pixel right along the columns. We consider a magnification factor $\alpha = 105$, so that the magnified image is shifted by an amount $(1+\alpha)\delta_2 = 106$ pixels along the columns. Again, We show the reference frame, the shifted frame and the magnified one. All three images have a resolution of $512 \times 512$ pixels, showing that the algorithm fits well also when $N$ and $M$ are both even numbers.\\
        In figure \ref{fig:AmpliTwoImag3}, $\delta_1 = 3$ and $\delta_2 = 4$, that is, all the intensity values of the second image are translated by 3 pixels down along the rows and 4 pixels left along the columns. We consider a magnification factor $\alpha = 25$, so that the magnified image is shifted by an amount $(1+\alpha)\delta_1 = 78$ pixels along the rows and $(1+\alpha)\delta_2 = 104$ pixels along the columns. The reference frame, shows a image with many circles in the scene, but all of these are subjected to the same amount of motion. All three images have a resolution of $1023 \times 1023$ pixels.
        
\begin{figure}[htpb!]
            \centering
            \subfigure[]{\includegraphics[width=0.3\textwidth]{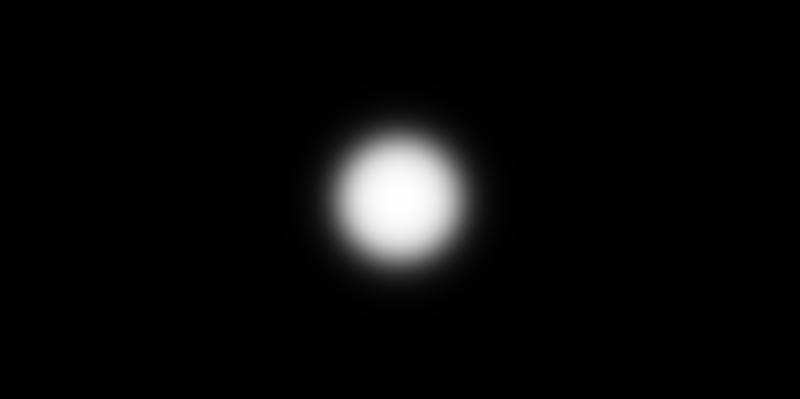}}
            \subfigure[]{\includegraphics[width=0.3\textwidth]{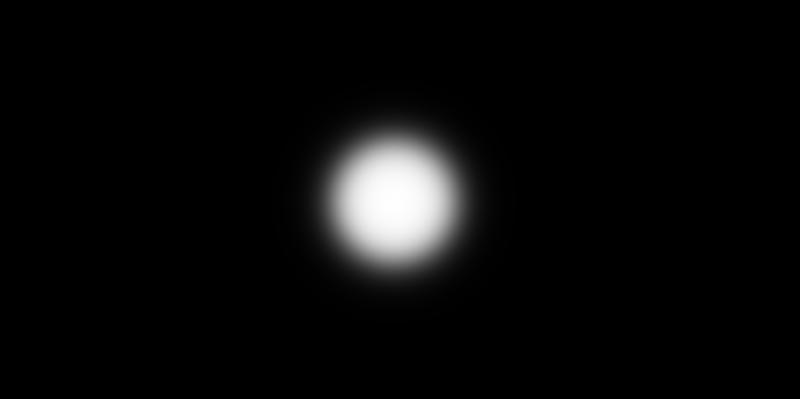}}
            \subfigure[]{\includegraphics[width=0.3\textwidth]{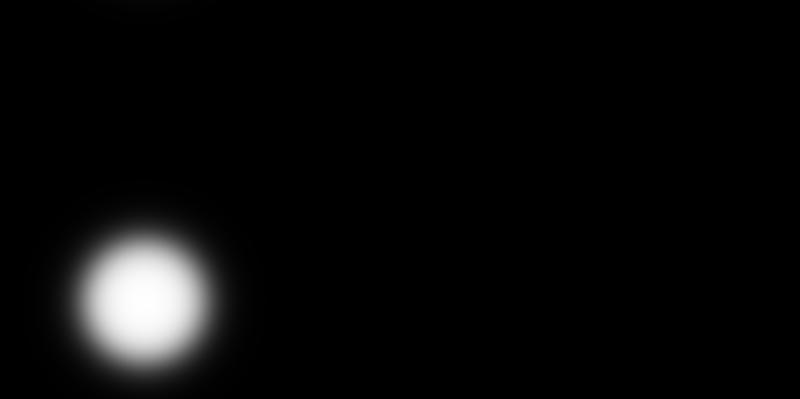}}
    
            \caption{(a) Original Frame (b) Shifted Frame (c) Magnified Frame ($\alpha = 50$)}
            \label{fig:AmpliTwoImag4}
        \end{figure}

        The procedure is effective in handling no squared images, and for images in which $N$ is odd and $M$ is even and vice versa. In fact, in Figure \ref{fig:AmpliTwoImag4}, we set $N = 399$ and $M = 800$, moreover we apply a shift of $\delta_1 = 2$ and $\delta_2 = -5$.This means that the second image is shifted by a 2 pixels down along the rows and 5 pixel left along the columns with respect to the reference one. The magnification factor used is $\alpha = 50$, so that the magnified image is shifted by an amount $(1+\alpha)\delta_1 = 102$ pixels along the rows and $(1+\alpha)\delta_2 = -255$ pixels along the columns.

        \subsection{Video Sequence}
         The following numerical example demonstrates the motion magnification process applied to a video sequence. The video consists of a series of frames of circles, generated in MATLAB following the method outlined previously. Specifically, the video spans $T = 5$ seconds, at a frame rate of $150$ frames per second. Every frame has a resolution of $709 \times 709$. Motion is introduced by applying time-varying displacements $\delta_1(t)$ and $\delta_2(t)$ to each time. More in detail the functions used to generate motion in the video sequence are:
         \begin{equation*}
             \begin{split}
                 \delta_1(t) &= \frac{1}{2}e^{-\frac{1}{2}t}\sin(2\pi \, 10 t),\\
                 \delta_2(t) &= \frac{1}{2}e^{-\frac{1}{2}t}\cos(2\pi \, 10t),
             \end{split}
         \end{equation*}
         where $t \in [0,5]$.\\
         We then apply the motion magnification procedure to the video sequence, amplifying the displacement between each consecutive frame with a magnification factor $\alpha = 40$. Notice that the shifting function are both such that $\delta_1(t), \delta_2(t) \leq \frac{1}{2}$ for $t \in [0,5]$ meaning that a sub-pixel motion is applied. \\
        \begin{figure}[htpb!]
				\includegraphics[width=6cm]{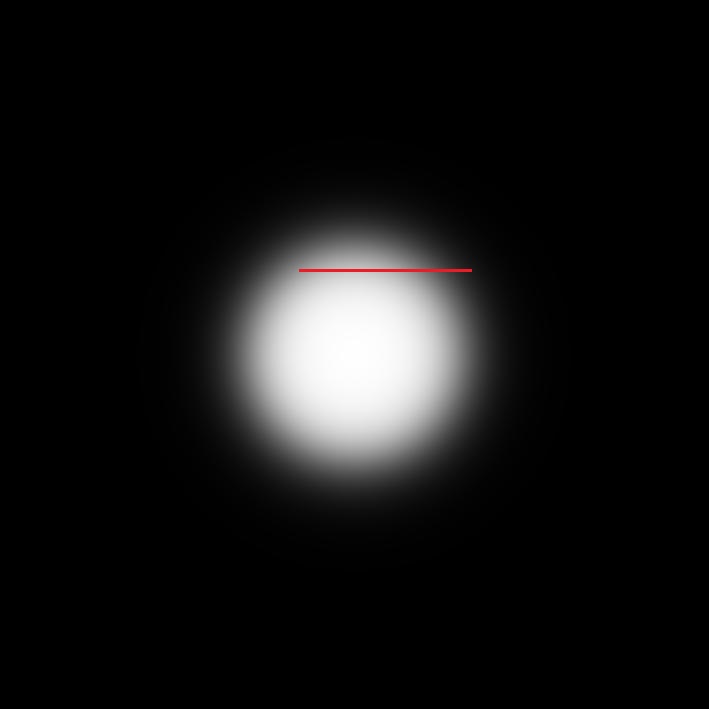}
				\centering
				\caption{The figure shows a frame of the video sequence in which we point out the line of intensities pixels value in red, used to exhibit the effect of the amplification.}
				\label{fig:OneLine}
	\end{figure}
        \begin{figure}[htpb!]
				\includegraphics[width=12cm]{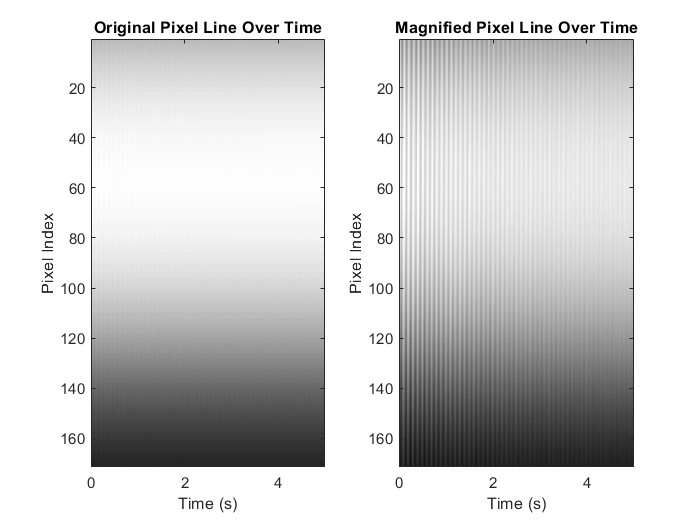}
				\centering
				\caption{A line of pixel values over time. The first plot shows the line of the original frame, the second one the line of the magnified video sequence.}
				\label{fig:LineOverTime}
	\end{figure}
         Figure \ref{fig:LineOverTime}, illustrates the effect of the magnification of this sub-pixel displacement. We plot the values of a pixel along a specific line (highlighted in red in Figure \ref{fig:OneLine}) as they vary over time. The plot on the left shows the pixel line over time without magnification, while the plot on the right displays the magnified version. As seen, the right image appears significantly more dynamic, demonstrating the enhanced motion achieved through the algorithm. The generated video frames can be viewed at \url{https://github.com/eTrebo98/MotMagArt1}.
	\section{Conclusion}
        In this paper, we show some preliminary results on the so called Motion Magnification algorithm. The goal of MM is to reveal subtle motion in a video sequence. Our procedure, which works in the Fourier domain, rely on the Fourier Shift Theorem, since motion in the original domain is related to a modification on the phase in the frequency one. That is, by amplifying the phase variation between two consecutive frame of a video we can obtain an output video in which the motion is exaggerated. This algorithm is effective for global motion, by the definition of the Fourier basis. The magnification procedure has been implemented in MATLAB and the current code version (v. 1.0) is available on Github. Future outline of work, surely involves to consider real video sequences, instead of the synthetic one. Moreover, since MM has applications in Vibration Analysis, extracting information on the shift $(\delta_1, \delta_2)$ has to be considered. Moreover, in general, in a video sequence, motions are not global, but local. That is, the shift depends both on the spatial coordinates and the temporal one. From this, the necessity to find adaptive procedures, which allows to extract and magnify local motions. For these methods, the usage of the Windowed Fourier Transform, or the Wavelet Transform can be helpful, since provides a localization of the image. Last but not least, instead on focusing on the phase information, amplitude-based MM techniques should be analyzed. In such one, the magnification is performed by directly modifying the intensity of the pixel values.
	
	\section*{Declaration of interests}
	The authors declare the following which may be considered as potential competing interests: first author's Erd\H{o}s number is 3 at the moment and, while getting 2 is still feasible, he can only cheat like this to get 1.
	
	\bibliography{jas_template_bib.bib}
	
	\newpage
	\appendix	
	\section{Declaration of interests} \label{app:dec}
	For the Declaration of interests, the authors must choose one of the following formulas, providing further details if necessary:
		\begin{itemize}
			
			\item[$\square$] The authors declare that they have no known competing financial interests or personal relationships that could have appeared to influence the work reported in this paper.
			
			\item[$\square$] The authors declare the following financial interests/personal relationships which may be considered as potential competing interests: $\ldots$ 
			
		\end{itemize}
	
	\section{Bibliography with Bib\TeX} \label{app:bib}
	\begin{itemize}
		
		\item References need to be provided in a .bib BibTeX database (best if exported from a big bibliographic database, such as \href{https://mathscinet.ams.org/mathscinet/index.html}{MathSciNet}).
		
		\item All references should be made with the command \verb*|\cite|, e.g,
		
		\item DOI and URL of the references should be included where available.
		
	\end{itemize}
	
\end{document}